# A Hybrid Decision Support System : Application on Healthcare

Abdelhak Mansoul, Baghdad Atmani, Sofia Benbelkacem

Department of Computer Science, Oran University, Algeria

mansoul.abdelhak@yahoo.fr, atmani.baghdad@gmail.com,
sofia.benbelkacem@gmail.com

## ABSTRACT

*Many systems based on knowledge, especially expert systems for medical decision support have been developed. Only systems are based on production rules, and cannot learn and evolve only by updating them. In addition, taking into account several criteria induces an exorbitant number of rules to be injected into the system. It becomes difficult to translate medical knowledge or a support decision as a simple rule. Moreover, reasoning based on generic cases became classic and can even reduce the range of possible solutions. To remedy that, we propose an approach based on using a multi-criteria decision guided by a case-based reasoning (CBR) approach.*

## 1. INTRODUCTION

In medicine, as in other areas, numerous applications in artificial intelligence have permitted the development of systems based on knowledge and particularly expert systems.

However, although this line of research has given rise to many scientific publications, expert systems routinely used are rare. And it became imperative to review traditional approaches of knowledge processing to propose solutions, and review the medical decision as a whole to reconsider the problem of decision support by a hybrid approach. Thus, it became possible to develop systems focused on medical action allowing clinicians to benefit from the possibilities offered by information technology and advanced processing such as data mining methods to improve their knowledge decisions and control their activities [1, 2, 3, 4].

We will explain in our present article the conception of the system as follows: In Section 2, we give some notions and concepts of decision support. In Section 3, we will establish a state of the art of the medical decision and the use of CBR in this area and in Section 4, we develop the conception of the proposed system.

## 2. MEDICAL DECISION SUPPORT SYSTEM

The medical decision support tends to provide clinicians with useful information after describing the clinical situation of the patient, in order to help them for improving the quality of care.

Thus, we can help the clinician in many different ways. Overall, three types of systems can be distinguished according to mode of intervention in the decision process.

a) Indirect decision support systems or documentary assistance systems
b) Systems for automatic reminders
c) Consulting systems

## 2.1 Concepts and Definitions

**Decision support.** "The decision support is the activity that is supported on models clearly explained but not necessarily completely formalized, helps get answers to the questions asked by a intervener in a decision-making process ... "[5]. This decision support, often builds on methods such as statistics, operations research, multi-criteria methods, etc.

**Medical decision support system.** "Computer program whose purpose is to provide physicians with timely and useful information describing the clinical situation of the patient and appropriate knowledge of this situation, properly filtered and presented to improve the quality of care and patients' health "[6].

**Multi-criteria Decision support.** In the context of multi-criteria decision the purpose of the decision is formed by a set of actions or alternatives.

**Problematics of Multi-criteria Decision.** For Roy [5], the real problems can be formulated using the multi-criteria analysis methods into four basic formulations: problematic of choice, denoted $P\alpha$, problematic of sorting or assignment denoted $P\beta$, problematic of storage denoted $P\gamma$ and the problematic of description $P\delta$.

To apply these methods, we usually use the following steps:

a) Identify the overall goal of the process and the type of decision.
b) List of actions and potential solutions.
c) Identify the criteria to guide decision makers.
d) Vote each solution with respect to each criterion.
e) Aggregate these judgments to select the most satisfactory solution.

The difference between the methods of multi-criteria analysis is mainly in the way of making the last step (e) or in how to evaluate each solution based on the criterion.

## 3. STATE OF THE ARTS

Due to the large volume of generated data in healthcare organizations, it has become imperative to take into account the mass of medical data to improve medical practice and even improve the care practiced by physicians. The methods of data mining, especially the decision trees, neural networks [7, 8, 9, 10, 11], have been put to use by many studies which we list here are a few: Sivakumar [12] presented a method based on neural networks to classify subjects with diabetic retinopathy (common complications of diabetes). Sung and Seong Hyeon [4] recently conducted a study based on the construction of a hybrid method, combining data mining methods to help doctors make faster and more accurate disease classification of chest pain.

**CBR in the medical field.** The use of CBR method is widely used in medicine precisely because the reasoning used and which is close to the physician faced with a given pathological situation. Indeed, a physician uses the same approach in seeking a medical solution based on his memory to try to remember the cases already experienced, and beyond it can easily move to a similar situation and if possible compare at its present position.

In addition, this approach is entirely justified in areas where finding a solution is not always based on a structured algorithmic method, but rather stored knowledge that is the solution of an experience.

Many works on the CBR of medical decision support systems were conducted, [13], Marling et al. [9] presented an approach to decision support based on CBR for the management of diabetes in patients with type 1 diabetes, systems have been developed for cardiac diagnosis "PROTOS" [14], CASIMIR [15] for the treatment of breast cancer. This list is far to be exhausted but shows the diversity of CBR application scope.

Compared to the epidemic of asthma, much works have led particular to understand this disease, for example trying to get feedback from the recorded data periodically on general medical consultations for asthma [16].

Other works have been oriented to decision support for the management of this disease [17, 18], and systems were squarely created for the diagnosis of asthma, such as Adema, [19] and Proforma [16].

This shows the interest in improving the treatment management of asthmatic patients by providing physicians with computer aids to medical decision.

## 4. THE PROPOSED SYSTEM

In the medical field, the use of the CBR approach is very interesting because the core of the reasoning process shows a strong similarity to the clinical reasoning. Indeed, the doctor often tries to make the connection between the case and those already experienced in his practice, and it is precisely the principle of this method.

In addition, the physician is often helped by the knowledge that medical stores, this knowledge is often related to many areas: medical, drugs, dosages, side effects of drugs, etc..

We are based on two aspects: CBR and medical knowledge stored to provide a support to medical decision process. This process consists of three modules: Data mining (DM), CBR and Multi-criteria Decision Support (MCDS).

Thus, the combination of the techniques of data mining and decision support can provide relevant information from different sources that help in making good decisions. Figures 1 & 2 show the process in question.

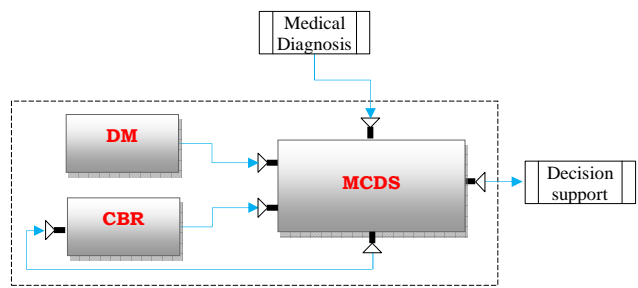

Figure 1. The adopted decision support system.

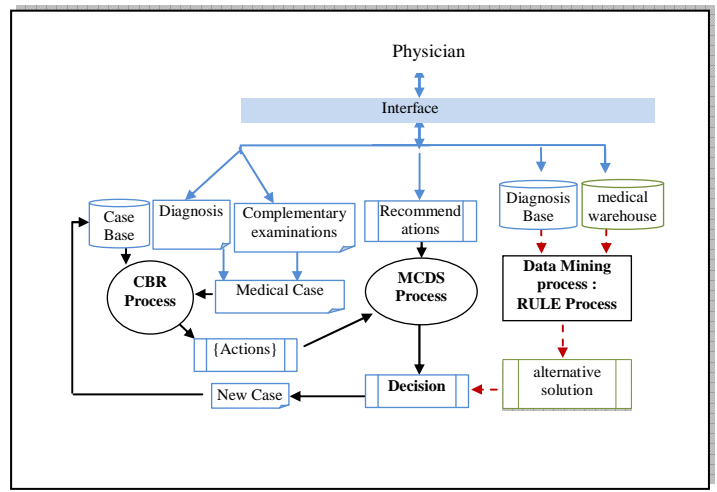

Figure 2. The proposed decision process.

## 4.1. The Proposed Medical Decision Support Process

### 4.1.1. The physician reasoning process

To describe a given pathological situation, the physician often uses his memory to search for two kinds of knowledge: expertise and cases already seen. Thus, if we want to formalize this process of reasoning and decision-making, we can write as follows:

**Physician reasoning (Expertise + cases already seen)**

This process can be translated into an automation and intelligence as a decision support, as follows:

**Decision support** (Knowledge Discovery + Use of scalable Case-base)

Thus, this process is necessarily a good and appropriate decision scheme which can easily be assimilated to a decisional model as follows:

*a)* Collect informations about the clinical case (diagnosis).

*b)* Consider a list of possible therapies.

*c)* Filter therapies.

*d)* Select the best therapy.

*e)* Review the choice of the proposed therapy.

*f)* Applying the decided therapy.

*g)* Check to confirm or reverse the decision of the chosen therapy.

### 4.1.2. The decisional model

We adopted the model of Simon (information, design, choice, review) because he is best suited to the decision scheme cited earlier (II), where we find the situation of the physician alone and facing a particular medical case.

### 4.1.3. The case-base

Our case is defined by a set of paraclinical descriptors such as sex, age, marital status ... etc., a set of clinical descriptors (symptoms) such as tension, fever ... etc., and a set of actions that have been effectively considered for the case in question.

Then we have:
$$Case\ (desc\_cli_1, desc\_cli_2, desc\_cli_3, ..., desc\_cli_n, desc\_parac_1, desc\_parac_2, ...., parac_m,$$
$$Action_1\_Case, Action_2\_Case, Action_3\_Case, ..., Action_p\_Case)$$

### 4.1.4. Reasoning through a case (CBR)

The CBR cycle to support the medical decision adopted (see Figure 3), is typically based on four tasks: retrieve, reuse, revise and retain. For the main task, retrieve, it is the search of the closest cases using a similarity measure. Then we use MVDM (Modified Value Difference Metric) [8], a similarity measure widely used to calculate the distance between nominal values (well suited to our medical descriptors describing the case).

This process intends to find all similar cases (SC) and all Action_Case (AC) that have been already produced in such situation to switch to the MCDS (Multi-Criteria Decision Support), to integrate them and support the conception stage process (analysis, aggregation, ... etc..) to pick the best actions for the case who is being processed.

### 4.1.4. The multi-criteria decision support (MCDS)

Our system is working on a problematic of selecting a subset as small as possible actions for a ultimate choice of one. This problematic is perfectly placed in front of the choice of therapy. For this, we use the Electra I method proposed by Roy [5] and solves the multi-choice problems by identifying the subset of actions with the best possible compromise.

MCDS will work on all of the nearest cases proposed by CBR process, exploiting their Action_Case in order to choose the best possible actions that will then be proposed in the solution of decision support.

### 4.2. The Field of Application

Diabetes is an incurable disease that occurs when the body is unable to properly use sugar (glucose), which is a "fuel" essential to its operation. Given this situation, we believe that we should try to support the effort of the management of this epidemic by physicians by providing a system or model that allows them to improve the quality of care that they provide to diabetic patients (children, adults or elderly).

Our study is intended to experiment with a multiple criteria decision approach to medical care in the diagnosis and the proposed therapy for diabetic patients.

To show the judicious choice of using ELECTRA I we use a classic and simplified example: a physician front of a pathological state (clinical case), and review the two main steps, namely information and design.

Example (fictitious).

After a medical exam, a physician found the following facts about a patient: excessive urination (it is frequent to getting up at night to urinate), increased thirst and hunger, weight loss, weakness and excessive fatigue, and blurred vision.

**Information.** It is the step of construction and representation of the case. The physician defines a pathological situation with a set of information (male / female, age, excessive urination, increased thirst and hunger, weight loss, etc...). This information will help to describe the situation or case. We will write then:

*Case* (Excessive urination, increased thirst and hunger, weight loss, weakness and excessive fatigue, blurred vision, planned actions)

**Conception**
- **Definition of the problematic and choice of method.**

Given the diversity of existing multi-criteria methods, we must select the one that can resolve the proposed case, in our case ELECTREA I is best suited because it addresses the problematic of choice of therapy (problematic α) in presence of several criteria which are all determinants.

- **Implementation of the ELECTRA I method**

The ELECTRA I application requires preliminary work before operating and that is to define the set of actions envisaged (therapies), the criteria and corresponding weights.

To assess and implement a therapy, allergist stands before him previously mentioned criteria for the 2 therapies considered: acting insulin for basal, rapid-acting insulin for bolus. From there, we have a Multi-Criteria Problem defined as follows: MCP (A, C, P).

A=therapy {acting insulin for basal, rapid-acting insulin for bolus}

C=Criterion {{side effects, {Many, No, Not at all}}, {treatment efficacy, {Very good, Good, Fair}}, {Duration of therapy, {long, reduced}}}

P = Weighting of therapy {{3, most appropriate treatment}, {2, least appropriate treatment},{1, treatment totally unsuitable}}

## 5. SYSTEM ARCHITECTURE

This is an Interactive Support System for Medical Decisions (ISS-MD) defined as a complete process, which includes a set of procedures to ensure the various features

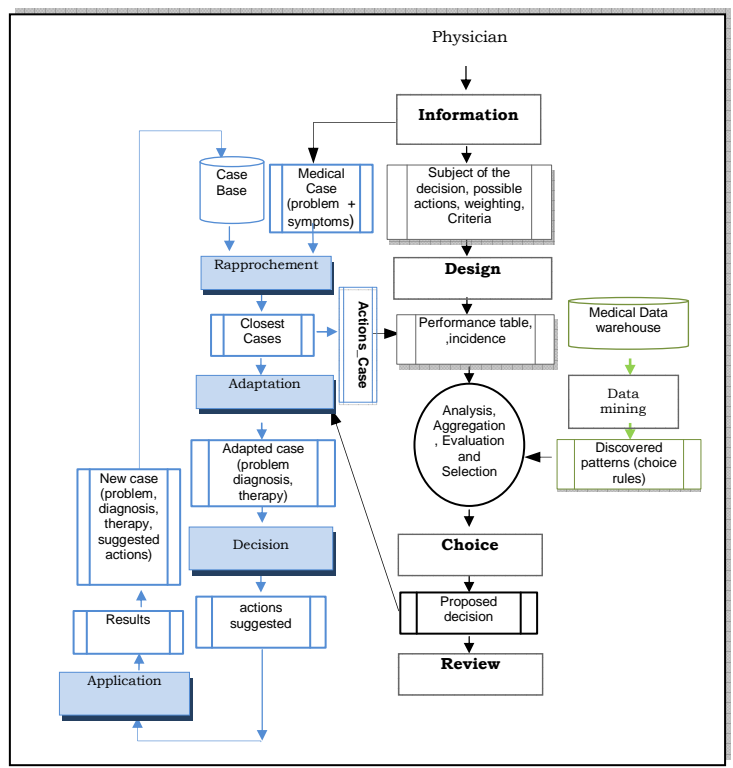

Figure 3. Architecture of the medical multi-criteria decision support.

### 5.1. The Decisional Process

The ISS-MD is defined as a complete processing chain which provides seven major process steps:

1) **Knowledge discovery.** By using an appropriate data mining method to mine medical interesting patterns (interesting rules for selecting actions) that help the MCDS process in the a priori definition of actions that can be considered.

2) **Information.** The physician define his clinical cases by priority and relevant information (objective and priorities) such as antecedents, clinical signs, etc. Then he defined the actions (therapy) that are deemed possible and finally identified and judge the evaluation criteria, preferences, and weights for these actions.

3) **Generation of rules for action choices and their criteria :** similar cases are selected from the case base and placed in a collection that will be used to extract the preliminary actions (Actions_Case) that have already been proposed before, to be appended to the possible actions that the physician has already defined (step 2 : information).

4) **Design.** Multi-criteria analysis and generation of possible actions by ELECTRA I method and followed by optimization, development, evaluation and selection of actions involved in decision support.

5) **Choice.** The physician will choose between different possible actions suggested to him and will decide to take them into consideration or not. This will allow the system to consider or not the new case.

6) **Review.** A review can be useful to refine the decision support by the physician.

## 6. EXPERIMENTATION

The purpose of the proposed approach is twofold: First, we start with building the training sample Case-base and then proceeds to decision support. For this we will use a medical database on diagnostic of diabetes, the Pima Indian diabetes database. It is a collection of medical diagnostic reports of 768 examples from a population living near Phoenix, Arizona, USA. The samples consist of examples with 9 attribute values and the last indicates one of the two possible outcomes, namely whether the patient is tested positive for diabetes. The database in the repository has 512 examples in the training set and 256 examples in the test set.

### 6.1. Pima+Indians+Diabetes Data Base

Each patient is represented in the data set by nine attributes as follows (in this order): Number of times pregnant, Plasma glucose concentration a 2 hours in an oral glucose tolerance test, Diastolic blood pressure (mm Hg), Triceps skin fold thickness (mm), 2-Hour serum insulin (mu U/ml), Body mass index (weight in kg/(height in m)^2), Diabetes pedigree function, Age (years). Finally, we have the ninth attribute Class variable (0 or 1) shows the diagnosis.

The figure below shows the structure of the database.

```
6,148,72,35,0,33.6,0.627,50,1
1,85,66,29,0,26.6,0.351,31,0
8,183,64,0,0,23.3,0.672,32,1
...............
```

Figure 4. Sample of Pima+Indians+Diabetes database*

* http://archive.ics.uci.edu/ml/datasets/Pima+Indians+Diabetes

## 6.2. Building of the training sample Case-base

Let $\Omega = \{\omega_1, \omega_2, ..., \omega_n\}$ training sample, this is the case set that will be used to build the case-base. Each case is described by a set of variables $X_1, X_2, ..., X_p$ called descriptive variables. for each case $\omega_i$ we associate a target attribute denoted Y which takes its values in the set of Diagnosis $Y = \{Y_1, Y_2, ....Y_k\}$.

Suppose that the training sample $\Omega$ obtained from the database Pima+Indians+Diabetes it contains number of cases $\omega_i$ described by 8 descriptive variables $X_1, X_2, ....., X_8$ and which is associated with a class $Y$ matching a diagnosis.

$X_1$ : Number of times pregnant

$X_2$: Plasma glucose concentration a 2 hours in an oral glucose tolerance test

$X_3$: Diastolic blood pressure (mm Hg)

$X_4$: Triceps skin fold thickness (mm)

$X_5$: 2-Hour serum insulin (mu U/ml)

$X_6$: Body mass index (weight in kg/(height in m)^2)

$X_7$ : Diabetes pedigree function

$X_8$ : Age (years)

Y : Diagnosis variable (0 or 1) = diagnosis

The following table (Table 1) shows a few cases from Pima+Indians+Diabetes database.

Table 1. Conversion of training sample $\Omega$ obtained from Pima+Indians+Diabetes database to a case base.

| $\omega$ | $X_1(\omega)$ | $X_2(\omega)$ | $X_3(\omega)$ | $X_4(\omega)$ | $X_5(\omega)$ | $X_6(\omega)$ | $X_7(\omega)$ | $X_8(\omega)$ | $Y(\omega)$ |
|---|---|---|---|---|---|---|---|---|---|
| $\omega_1$ | 6 | 148 | 72 | 35 | 0 | 33.6 | 0.627 | 50 | 1 |
| $\omega_2$ | 1 | 85 | 66 | 29 | 0 | 26.6 | 0.351 | 31 | 0 |
| ... | ... | ... | ... | ... | ... | ... | ... | ... | ... |
| ... | | | | | | | | | |
| ... | | | | | | | | | |
| $\omega_6$ | 5 | 116 | 74 | 0 | 0 | 25.6 | 0.201 | 30 | 0 |
| ... | | | | | | | | | |
| ... | | | | | | | | | |
| ... | | | | | | | | | |

In this example, Y belongs to the set of Diagnosis Y= {O, 1}, where 0 = "tested negative for diabetes" and 1 = "tested positive for diabetes"

The new system is developed in JAVA with an interconnecting module to the JCOLIBRI system [20]. This system is essentially based on an engine described by the following algorithm MCDS.

We use Jcolibri platform for building the case-base, then we'll get the result of this platform and give it to MCDS (developed in Java) to produce decision support. The purpose of this model is to decide the diagnosis (assign a class) to each new case given as input.

According to the previous description of the proposed model, whole process will be done by the MCDS pseudo-algorithm:

```
Algorithm : MCDS (RBC+Multi-Criteria)
    Input : Medical_case (Problem, symptoms, subject, possible_actions) / description of case
    Output : Actions_suggested
    Begin.
    Information (Object_Decision,weight,Criteria)
    Define_Case (Problem,Symptomes,Possible_Actions,CASE)
    RBC.Rapprochement (CASE, Result_Rapprochement)
    If  Result_Rapprochement=Set of cases                        / the case exist
        For each  CASE  in Set of cases
            {Actions_Case=Actions_Case+Current_Actions_Case()}
        Endfor
    Else
        Actions_Case= Ø                                          / new Case
    EndIf
    Actions_Case=Actions_Case+Possible_Actions
    Choice_Rules=DM()
    Performance_Table, Actions_Case, Incidences=Electra_1 ()
    Electra_1_AAES ( Actions_Case, Performance_Table, Incidences, Choice_Rule)
    Proposed_Decision=Choice()
    Proposed_Decision=Review(Proposed_Decision)
    RBC.Adaptation(Proposed_Decision,Adapted_Case)
    Actions_Suggested=RBC.Decision( )
    RBC.Application(Actions_Suggested,Results, new_case)
    Storage_new_case(new_case)
    End.
```

### 6.3. Results of the experimentation

To evaluate the efficiency of our approach, we tested it on a Pima+Indians+Diabetes database that we transformed into case-base.

There is an important attribute (attribute 2: Plasma glucose concentration a 2 hours in an oral glucose tolerance test). Only trying to change its value we can already tipping the case to a positive or negative diagnosis.

Then we can enter values for other attributes and ask the system to make a decision aid.

To perform such tests we introduced values for 10 cases supposed to be positive diagnosis and 10 cases with the assumption that they are negative diagnosis.

A comparison of each case introduced is then made with the real case-base. And the system gives results.

We calculate the rate of positive (%) and the rate of negative (%) cases found based on the introduced cases. This rate represents the number of cases found in the case-base and reported according as they are introduced. The results are presented in Table 3.

From the results, we note that the rate of positive and negative is more than the average which indicates that our system tend to give answers to reality as declared in the case base, especially of positive cases.

This result is summarized in the following table:

Table 2. Results of the experimentation.

| Diagnosis for diabetes | Case Diagnosis | Number of cases | Introduced cases supposed + | Introduced cases supposed - | (%) Result + | (%) Result – |
|---|---|---|---|---|---|---|
| Negatif | 0 | 500 | | 10 | | 50 |
| Positif | 1 | 268 | 10 | | 60 | |

## 7. CONCLUSION

This present study provides the theoretical basis of an approach that tends to solve a problematic of decision support. This approach is based on case-based reasoning and multi-criteria. These tools are well adapted to the medical context. We aim to develop a new generation of decision support techniques that use multiple tools called hybrid decision support systems.

The designed MCDS facilitates the optimization of action choices, a complete and well-integrated process, to help and guide all phases of the decision. In a later step we intend to develop by enriching different multi-criteria methods to solve the problematic of sorting and storage based on clinical situations that may occur to the physician, so that he can provide a way for himself to model the problem (clinical situation) in different ways.

On another axis, we intend adding to our model various therapy schemes for different types of diabetes (Type 1 Diabetes, Type 2 Diabetes, gestational Diabetes, etc.) to help refinement of decision support by typical therapies.

**Authors**

**Abdelhak MANSOUL** is an Assistant Professor at Skikda University and affiliated researcher in Oran Computer Lab of Oran University.
His research interests are in Database Management System, Data Mining, decision support systems, and simulation.

**Baghdad ATMANI** received his Ph.D. degree in computer science from the University of Oran (Algeria), in 2007. His interest field is Data Mining and Machine Learning Tools. His research is based on Knowledge Representation, Knowledge-based Systems and CBR, Data and Information Integration and Modeling, Data Mining Algorithms, Expert Systems and Decision Support Systems.
His research are guided and evaluated through various applications in the field of control systems, scheduling,

production, maintenance, information retrieval, simulation, data integration and spatial data mining

**Sofia BENBELKACEM** is currently a Ph.D. candidate in the Computer Science Department at the University of Oran, and affiliated researcher in Oran Computer Lab. Her research interests include Data Mining, planning, case-based reasoning and medical decision support systems.